\documentclass{tlp}

\usepackage{aopmath}

\newcommand{\barp}{\ensuremath{I_P}}
\newtheorem{algorithm}{Algorithm}

\begin{document}

\bibliographystyle{acmtrans}

\title[Intelligent Backtracking] 
{Enhancing a Search Algorithm to Perform Intelligent Backtracking} 

\author[Maurice Bruynooghe]
{MAURICE BRUYNOOGHE \\
Katholieke Universiteit Leuven, Department of Computer
  Science
\\Celestijnenlaan 200A, B3001 Heverlee, Belgium\\
 e-mail: Maurice.Bruynooghe@cs.kuleuven.ac.be
}
\pagerange{\pageref{firstpage}--\pageref{lastpage}}
\volume{\textbf{1} (1):}
\jdate{January 2001}
\setcounter{page}{1}
\pubyear{2001}

\maketitle[p]

\shorttitle{Programming pearl}

\label{firstpage}

\begin{abstract}
  This paper illustrates how a Prolog program, using chronological
  backtracking to find a solution in some search space, can be
  enhanced to perform intelligent backtracking. The enhancement
  crucially relies on the impurity of Prolog that allows a program to
  store information when a dead end is reached. To illustrate the
  technique, a simple search program is enhanced.

To appear in Theory and Practice of Logic Programming.

\end{abstract}

\begin{keywords}
 intelligent backtracking, dependency-directed
  backtracking, backjumping, conflict-directed backjumping, nogood
  sets, look-back.  
\end{keywords}

\section{Introduction}
\label{sec:intro}

The performance of backtracking algorithms for solving finite-domain
constraint satisfaction problems can be improved substantially by so
called look-back and look-ahead methods \cite{Dechter02}. Look-back
techniques extract information by analyzing failing search paths that
are terminated by dead ends and use that information to prune the
search tree. Look-ahead techniques use constraint propagation
algorithms in an attempt to avoid such dead ends altogether.
Constraint propagation can rather easily be isolated from the search
itself and can be localized in a constraint store. Following the
seminal work of \cite{Hentbook}, look-ahead techniques are available
to the logic programmer in a large number of systems.

This is not the case for look-back methods.  Intelligent backtracking
has been explored as a way of improving the backtracking behavior of
logic programs \cite{BP84}.  For some time, a lot of effort went into
adding intelligent backtracking to Prolog implementations (see
references in \cite{Br91}). However, the inherent space and time
costs, which must be paid even when no backtracking occurs, impeded
its introduction in real implementations.

For a long time, look-ahead methods dominated in solving constraint
satisfaction problems. However, already in \cite{RB86} we have shown
empirical evidence that look-back methods can be useful, even that it
can be interesting to combine both. Starting in the nineties there is
a renewed interest in look-back methods, {\em e.g.}, \cite{ginsberg93},
and in combining look-back with look-ahead {\em e.g.},
\cite{Dechter02}.

Look-back turned out to be the most successful of the approaches tried 
in a research project aiming at detecting unsolvable queries (queries
that do not terminate, such as the query $\leftarrow \mathit{ odd}(X),
\mathit{ even}(X)$ for a program defining odd and even numbers). The
approach was to construct a model of the program over a finite domain
in which the query was false. The central part of this model
construction was to search for a pre-interpretation leading to the
desired model, {\em i.e.}, with $D$ the domain, to find an appropriate
function $D^n \rightarrow D$ for every n-ary functor in the program. A
meta-interpreter was built which performed a backtracking search over
the solution space. A control strategy was devised which resulted in
the early detection of instances of program clauses which showed that
the choices made so far could not result in the desired model. This
meta-interpreter outperformed dedicated model generators on several
problems \cite{BVWD98}. However it remained very sensitive to the
initial ordering in which the various components of the different
functions were assigned. The point was that not all choices made so
far necessarily contributed to the evaluation of a clause instance.
We experimented with constraint techniques and also investigated the
use of intelligent backtracking. With a small programming effort, we
could enhance the meta-interpreter to support a form of intelligent
backtracking. As reported in \cite{BVWD99}, this was the most
successful approach. As Prolog is a popular tool for prototyping
search problems and as look-back methods, though useful, are not
available in off-the-shelf Prolog systems, we decided to describe for
a wider audience how to enhance a Prolog search program with a form of
intelligent backtracking. The technique crucially depends on the
impure feature of Prolog (assert/retract) that allows storing
information when a dead end is reached. The stored information is used
to decide whether a choice point should be skipped when chronological
backtracking returns to it. Hence we propose the technique as a black
pearl.

In the application mentioned above, the meta-interpreter is performing
a substantial amount of computation after making a choice whereas the
amount of computation added to support intelligent backtracking is
comparatively small. This is not always the case. When the amount of
computation in between choices is small and solutions are rather easy
to find, the overhead of supporting intelligent backtracking may be
larger than the savings due to the pruning of the search space. This
is the case in toy problems such as the n-queens. In the example we
develop here, there is a small speed-up.

We recall some basics of intelligent
backtracking in Section~\ref{sec:IB}. In Section~\ref{sec:nqueens}, we
introduce the example program and in Section \ref{sec:ib} we enhance
it with intelligent backtracking. We conclude with a discussion in
Section~\ref{sec:discussion}. 

\section{Intelligent Backtracking}
\label{sec:IB}

Intelligent backtracking as described in \cite{Br81} is a very general
schema. It keeps track of the reason for eliminating a variable in a
domain. Upon reaching a dead end, it identifies a culprit for the
failure and {\em jumps back} to the choice point where the culprit was
assigned a value. Information about the variables assigned in between
the culprit and the dead end can be retained if still valid, as in the
dynamic backtracking of \cite{ginsberg93} which can be considered as
an instance of the schema. More straightforward in a Prolog
implementation is to give up that information, this gives the
backjumping algorithm (Algorithm 3.3) in \cite{ginsberg93}
(intelligent backtracking with static order in \cite{RB86}). We follow
rather closely \cite{ginsberg93} for introducing it.

A constraint satisfaction problem (CSP) can be identified by a triple
$(I,D,C)$ with $I$ a set of variables, $D$ a mapping from variables to
domains and $C$ a set of constraints. Each variable $i \in I$ is
mapped by $D$ into a domain $D_i$ of possible values.  Each constraint
$c \in C$ defines a relation $R_c$ over a set $I_c \subseteq I$ of
variables and is satisfied for the tuples in that relation.  A
solution to a CSP consists of a value $v_i$ (an {\em assignment}) for
each variable $i$ in $I$ such that: (1) for all variables $i$: $v_i
\in D_i$ and, (2) for all constraints $c$: with $I_c = \{j_1, \ldots,
j_k\} $, it holds that $(v_{j_1}, \ldots, v_{j_k}) \in R_c $.

A partial solution to a CSP $(I,D,C)$ is a subset $J \subseteq I$ and
an assignment to each variable in $J$. A partial solution $P$ is
ordered by the order in which the algorithm that computes it assigns
values to the variables and is denoted by a sequence of ordered pairs
$(i,v_i)$. A pair $(i,v_i)$ indicates that variable $i$ is assigned
value $v_i$; $\barp = \{i | (i,v_i) \in P\}$ denotes the set of
variables assigned values by $P$.

Given a partial solution $P$, an {\em eliminating explanation}
(cause-list in \cite{Br81}) for a variable $i$ is a pair $(v_i,S)$
where $v_i \in D_i$ and $S \subseteq \barp$. It expresses that the
assignments to the variables of $S$ by the partial solution $P$ cannot
be extended into a solution where variable $i$ is assigned value
$v_i$.  Contrary to \cite{ginsberg93}, we use an {\em elimination
  mechanism} that tests one value at a time. Hence we assume a
function $\mathit{ consistent}(P, i ,v_i)$ that returns true when $P
\cup \{(i,v_i)\}$ satisfies all constraints over $\barp \cup \{i\})$
and a function $\mathit{ elim}(P, i ,v_i)$ that returns an eliminating
explanation $(v_i,S)$ when $\neg\mathit{ consistent}(P, i ,v_i)$.

Below, we formulate the backjumping algorithm; next we clarify its
reasoning.  $E_i$ is the set of eliminating explanations for
variable $i$.

\begin{algorithm}
\label{alg:1}
  Given as inputs a CSP $(I,D,C)$.
  \begin{enumerate}
  \item Set $P:=\emptyset$.
  \item If $\barp = I$ return $P$. Otherwise select a variable $i \in I
    \setminus \barp$, set $S_i := D_i$ and $E_i := \emptyset$.
  \item If $S_i$ is empty then go to step 4; otherwise, remove an element
    $v_i$ from it.\\ 
    If $\mathit{ consistent}(P, i ,v_i)$ then extend $P$ with
    $(i,v_i)$ and go to step 2; otherwise add $\mathit{ elim}(P, i
    ,v_i)$ to $E_i$ and go to step 3.
  \item ($S_i$ is empty and $E_i$ has an eliminating explanation for
    each value in $D_i$.) Let $C$ be the set of all variables appearing
    in the explanations of $E_i$.
  \item If $C = \emptyset$, return failure. Otherwise, let $(l,v_l)$
    be the last pair in $P$ such that $l \in C$. Remove from $P$ this
    pair and any pair following it. Add $(v_l,C \setminus \{l\})$ to
    $E_l$, set $i:=l$ and go to step 3.
  \end{enumerate}
\end{algorithm}

In step 3, when the extension of the partial solution is inconsistent
then $\mathit{ elim}(P, i ,v_i)$ returns a pair
$(v_i,\{j_1,\ldots,j_m\})$ such that the partial solution
$(j_1,v_{j_1}), \ldots ,(j_m,v_{j_m}),(i,v_i) $ violates the
constraints. The inconsistency of this assignment can be expressed by
the clause: $\leftarrow j_1 = v_{j_1}, \ldots ,j_m=v_{j_m}, i=v_i
$ (The head is false, the body is a conjunction).

In step 4, when $S_i$ is empty, we have an eliminating explanation for
each value $v_{i_k}$ in the domain $D_i$. Hence we have a set of
clauses of the form
\begin{equation}
  \label{cl2}
  \leftarrow j_{k,1} = v_{j_{k,1}}, \ldots, 
  j_{k,m_k} = v_{j_{k,m_k}}, i=v_{i_k}
\end{equation}

The condition that the variable $i$ must be assigned a value from
domain $D_i$ with $n$ elements can be expressed by the clause (the
head is a disjunction, the body is true):
\begin{equation}
  \label{cl1}
  i =v_{i_1}  ,\ldots, i =v_{i_n} \leftarrow
\end{equation}

Now, one can perform hyperresolution \cite{Rob65} between clause
(\ref{cl1}) and the clauses of the form (\ref{cl2}) (for $k$ from 1 to
$n$). This gives:
\begin{equation}
  \label{cl4}
  \leftarrow j_{1,1} = v_{j_{1,1}}, \ldots, 
  j_{1,m_1} = v_{j_{1,m_1}}, \ldots,
  j_{n,1} = v_{j_{n,1}}, \ldots ,
  j_{n,m_n} = v_{j_{n,m_n}}
\end{equation}

This expresses a conflict between the current values of the variables
in the set $\{j_{1,1}, \ldots, j_{1,m_1}, \ldots, j_{n,1},
\ldots,j_{n,m_n} \} = C$. Hence, with $l$ the last assigned variable
in $C$, $C\setminus \{l\}$ is an eliminating explanation for
$v_l$.  The conflict $C$ is computed in step 4. When empty, the
problem has no solution as detected in step 5. Otherwise, step 5
backtracks and adds the eliminating explanation $(v_l, C\setminus
\{l\})$ to the set of eliminating explanations of variable $l$.

One can observe that the algorithm does not use the individual
eliminating explanations in the set $E_i= (v_{i_k}, S_k)$, but only
the set $C$ which is the union of the sets $S_k$. As we have no
interest in introducing more refined forms of intelligent
backtracking, we develop Algorithm~\ref{alg:2} where $E_i$ holds the
union of the sets $S_k$ in the eliminating explanations of variable
$i$. To obtain an algorithm that closely corresponds to the Prolog
encoding we present in Section~\ref{sec:ib}, we reorganise the code
and introduce some more changes. The function $\mathit{
  elim}(P,i,v_i)$ that returns an eliminating explanation $(v_i,S)$
for the current value of variable $i$ is replaced by a function
$\mathit{ conflict}(P,i,v_i)$ that returns the set $\{i\} \cup S$ (the
variables that participate in a conflict as represented by
Equation~\ref{cl2}). This conflict is stored in a variable $C$ (step 3
of Algorithm~\ref{alg:2}). It is nonempty and $i$ is the last assigned
variable, hence the value of $i$ remains unchanged in step 4 and, in
step 5, the eliminating explanation $C \setminus \{i\}$ is added to
$E_i$. This reorganisation of the code has as result that a local
conflict (the chosen value for the last assigned variable $i$ is
inconsistent with the partial solution) and a deep conflict (all
values for variable $i$ have been eliminated) are handled in a uniform
manner: upon failure, the algorithm computes a conflict and stores it
in variable $C$ (for the local conflict in step 3, for the deep
conflict in step 5), backtracks to the variable computed in step 4
(the ``culprit'') and resumes in step 5 with updating $E_i$ and trying
a next assignment to variable $i$.

\begin{algorithm}
\label{alg:2}
  Given as input a CSP $(I,D,C)$.
  \begin{enumerate}
  \item Set $P:=\emptyset$.
  \item If $\barp = I$ return P. Otherwise select a variable $i \in I
    \setminus \barp$. Select a value $v_i$ from $D_i$. Set $S_i := D_i
    \setminus \{v_i\}$ and $E_i := \emptyset$.
  \item If $\mathit{consistent}(P, i ,v_i)$ then extend $P$ with
    $(i,v_i)$ and go to step 2; otherwise set $C := \mathit{conflict}(P, i
    ,v_i)$. 
  \item If $C=\emptyset$ then return failure; otherwise let $(l,v_l)$
    be the last pair in $P$ such that $l \in C$. Set $i:=l$. 
  \item Add $C \setminus \{i\}$ to $E_i$. If $S_i= \emptyset$ then $C
    := E_i$ and go to step 4; otherwise select and remove a value
    $v_i$ from $S_i$ and go to step 3.
  \end{enumerate}
\end{algorithm}

\section{A search problem}
\label{sec:nqueens}

The code below is, apart from the specific constraints, fairly
representative for a finite domain constraint satisfaction problem.
The problem is parameterized with two cardinalities: {\tt VarCard},
the number of variables (the first argument of {\tt problem/3}) and
{\tt ValueCard}, the number of values in the domains of the variables
(the second argument of {\tt problem/3}). The third argument of {\tt
  problem/3} gives the solution in the form of a list of elements
$\mathit{ assign}(i,v_i)$. The main predicate uses {\tt
  init\_domain/2} to create a domain $[1, 2, \ldots,
\mathit{ValueCard}]$ and {\tt init\_pairs/3} to initialize
$\mathit{Pairlist}$ as a list of pairs $i$-$D_i$ with $D_i$ the domain
of variable $i$. The first argument of {\tt extend\_solution/3} is a
list of pairs $i$-$D_i$ with $i$ an unassigned variable and $D_i$ what
remains of its domain; the second argument is the (consistent) partial
solution (initialized as the empty list) and the third argument is the
solution. The predicate is recursive; each iteration extends the
partial solution with an assignment to the first variable on the list
of variables to be assigned. The nondeterministic predicate {\tt
  my\_assign/2} selects the value. If desirable, one could introduce a
selection function which dynamically selects the variable to be
assigned next.

Consistency of the new assignment with the partial solution is tested
by the predicates {\tt consistent1/2} and {\tt consistent2/2}. They
create a number of binary constraints. The binary constraints
themselves are tested with the predicates {\tt constraint1/2} and {\tt
  constraint2/2}. What they express is not so important. The purpose
is to create a problem that is sufficiently difficult so that
enhancing the program with intelligent backtracking pays off. For the
interested reader, the predicate {\tt consistent2/2} creates a very
simple constraint that verifies (using {\tt constraint1/2}) that the
value of the newly assigned variable is different from the value of
the previously assigned variable. The predicate {\tt consistent1/2}
creates a set of more involved constraints. The odd numbered and even
numbered variables each encode the constraints of the n-queens
problem. As a result, the solution of {\em e.g.,} {\tt problem(16,8,S)}
contains a solution for the 8-queens problem in the odd numbered
variables and a {\em different} (due to the constraints created by {\tt
  consistent2/2}) solution in the even numbered variables. Substantial
search is required to find a first solution.  For example, the first
solution for {\tt problem(16,8,S)} is found after 32936 assignments
(using a similar set-up of constraints, a solution is found for the
8-queen problem after only 876 assignments).

Note that the constraint checking between the new assigned variable
and the other assigned variables is done in an order that is in
accordance with the order of assigning variables. Hence {\tt
  consistent1/2} is not tail recursive.  The order is not important
for the algorithm without intelligent backtracking. However, it is
crucial to obtain optimal intelligent backtracking: as with
chronological backtracking, constraint checking will stop at the first
conflict detected and an eliminating explanation will be derived from
it. As an eliminating explanation with an older assigned variable
gives more pruning than one with a more recently assigned variable,
the creation of constraints requires one to pay attention to the
order. It is done already here to minimize the differences between
this version and the enhanced version.

\begin{verbatim} 
problem(VarCard,ValueCard,Solution) :-
        init_domain(ValueCard,Domain),
        init_pairs(VarCard,Domain,Pairs),
        extend_solution(Pairs,[],Solution).

init_domain(ValueCard,Domain) :-
        ( ValueCard=0 -> Domain=[]
        ; ValueCard>0, ValueCard1 is ValueCard-1, 
            Domain=[ValueCard|Domain1],
            init_domain(ValueCard1,Domain1)
        ).

init_pairs(VarCard,Domain,Vars) :-
        ( VarCard=0 -> Vars = []
        ; VarCard>0, VarCard1 is VarCard-1,
            Vars=[VarCard-Domain|Vars1],
            init_pairs(VarCard1,Domain,Vars1)
        ).

extend_solution([],Solution,Solution).
extend_solution([Var-Domain|Pairs],PartialSolution,Solution) :-
        my_assign(Domain,Value),
        consistent1(PartialSolution,assign(Var,Value)),
        consistent2(PartialSolution,assign(Var,Value)),
        extend_solution(Pairs,
                       [assign(Var,Value)|PartialSolution],
                       Solution).

my_assign([Value|_],Value).
my_assign([_|Domain],Value) :- my_assign(Domain,Value).

consistent1([],_).
consistent1([_],_).
consistent1([_, Assignment1|PartialSolution],Assignment0) :- 
        consistent1(PartialSolution,Assignment0),
        constraint1(Assignment0,Assignment1),
        constraint2(Assignment0,Assignment1).

consistent2([],_).
consistent2([Assignment1|_],Assignment0) :- 
        constraint1(Assignment0,Assignment1).

constraint1(assign(_,Value0),assign(_,Value1)) :- Value0 \== Value1.

constraint2(assign(Var0,Value0),assign(Var1,Value1)) :-
        D1 is abs(Value0-Value1),
        D2 is abs(Var0-Var1)//2,
        D1 \== D2.
\end{verbatim}

\section{Adding intelligent backtracking}
\label{sec:ib}

Adding intelligent backtracking requires us to maintain eliminating
explanations. In Algorithm~\ref{alg:2}, a single eliminating
explanation is associated with each variable. The eliminating
explanation of a variable $i$ is initialised as empty in step 2, when
assigning a first value to the variable. It is updated in step 5, when
the last assigned value turns out to be the ``culprit'' of an
inconsistency.  This happens just before assigning the next value to
variable $i$. This indicates that the right place to store eliminating
explanations is as an extra argument in the predicate {\tt
  my\_assign/2}. In step 4, the algorithm has to identify the ``last''
variable $l$ of a conflict (the ``culprit''), just before updating the
eliminating explanation. We will also use the {\tt my\_assign/2}
predicate to check whether the variable it assigns corresponds to the
culprit of the failure.  Hence also the identitity of the variable
should be an argument. These considerations lead to the replacement of
the {\tt my\_assign/2} predicate by the following {\tt my\_assign/4}
predicate.

\begin{verbatim}
my_assign([Value|_],_Var,_Explanation,Value ).
my_assign([_|Domain],Var,Explanation0,Value) :-
        get_conflict(Conflict),
        remove(Var,Conflict,Explanation1),
        set_union(Explanation0,Explanation1,Explanation),
        my_assign(Domain,Var,Explanation,Value).
my_assign([],_Var,Explanation,_Value) :-
        save_conflict(Explanation), fail.
\end{verbatim}

\noindent

It is called from {\tt extend\_solution/4} as {\tt
  myassign(Domain,Var,[],Value)} (what remains of the domain is the
first argument, the second argument is the variable being assigned,
the third argument is the initially empty eliminating explanation and
the fourth argument returns the assigned value). The initial call
together with the base case perform the otherwise branch of step 2.
The second clause, entered upon backtracking when the domain is
nonempty, checks whether the variable being assigned is the culprit.
To do so, it needs the conflict. As this information is computed just
before failure occurs, it cannot survive backtracking when using the
pure features of Prolog. One has to rely on the impure features for
asserting/updating clauses. Either {\tt assert/1} and {\tt retract/1}
or more efficient variants of specific Prolog systems\footnote{In our
  experiments, we made use of SICStus Prolog and employed {\tt
    bb\_put/2} and {\tt bb\_get/2}.}. The call to {\tt
  get\_conflict(Conflict)} picks up the saved conflict\footnote{We
  implemented it as {\tt get\_conflict(Conflict) :-
    bb\_get(conflict,Conflict)}.}; next, the call {\tt
  remove(Var,Conflict,Explanation1)} checks whether {\tt Var} is part
of it. If not, {\tt my\_assign/4} fails and backtracking returns to the
previous assignment. If {\tt Var} is the culprit, then the code
performs step 5 of the algorithm: {\tt remove/3} returns the
eliminating explanation in its third argument, {\tt set\_union/3} adds
it to the current eliminating explanation and the recursive call
checks whether the domain is empty. If not, the base case of {\tt
  my\_assign/4} assigns a new value. If the domain is empty, then the
last clause is selected. The eliminating explanation becomes the
conflict and is saved with the call to {\tt
  save\_conflict(Explanation)} that relies on the impure
features\footnote{We implemented it as {\tt save\_conflict(Conflict)
    :- bb\_put(conflict,Conflict)}.} and the clause fails.

Further modifications are in the predicates {\tt constraint1/2} and
{\tt constraint2/2} that perform the constraint checking. If a
constraint fails, the variables involved in it make up the conflict
and have to be saved so that after re-entering {\tt myassign/4} the
conflict can be picked up and used to compute an eliminating
explanation (step 3).  As the last assigned variable participates in all
constraints, it is part of the conflict.  For example, the code for
{\tt constraint1/2} becomes:

\begin{verbatim}
constraint1(assign(Var0,Value0),assign(Var1,Value1)) :-
        ( Value0 \== Value1 -> true
        ; save_conflict([Var0,Var1]), fail
        ).
\end{verbatim}

The modification to {\tt constraint2/2} is similar.  Recall that the
order in which constraints are checked determines the amount of
pruning that is achieved.  Finally, if one is interested in more than
one solution then also a conflict has to be stored when finding a
solution. It consists of all variables making up the solution. Using a
predicate {\tt allvars/2} that extracts the variables from a solution,
the desired behavior is obtained as follows:
\begin{verbatim}
problem(VarCard,ValueCard,Solution) :-
        init_domain(ValueCard,Domain),
        init_pairs(VarCard,Domain,Pairs),
        extend_solution(Pairs,[],Solution),
        initbacktracking(Solution).

initbacktracking(Solution) :-
        allvars(Solution,Conflict),
        save_conflict(Conflict).
\end{verbatim}

The enhanced program generates the same solutions as the original, and
in the same order. For {\tt problem(16,8,S)} the number of assignments
goes down from 32936 to 4015 and the execution time from 140ms to
70ms; for {\tt problem(20,10,S)}, the reduction is respectively from
75950 to 15813 and from 370ms to 310ms. The achieved pruning more than
compensates for the (substantial) overhead of recording and updating
conflicts\footnote{Using {\tt bb\_get} and {\tt bb\_put} to count the
  number of assignments increases execution time of the initial
  algorithm for {\tt problem(16,8,S)} from 140ms to 400ms.}  and of
the calls to {\tt remove/3} and {\tt set\_union/3}. Note that the
speed-up decreases with larger instances of this problem. This is
likely due to the increasing overhead of the latter two predicates.
Keeping the conflict set sorted (easy here because the variable
numbers corresponds with the order of assignment) such that the
culprit is always the first element could reduce that overhead.

\section{Discussion}
\label{sec:discussion}

In this black pearl, we have illustrated by a simple example how a
chronological backtracking algorithm can be enhanced to perform
intelligent backtracking. As argued in the introduction, look-back
techniques are useful in solving various search problems. Hence
exploring their application can be very worthwhile when building a
prototype solution for a problem. The technique presented here
illustrates how this can be realized with a small effort when
implementing a prototype in Prolog. Interestingly, the crucial feature
is the impurity of Prolog that allows the search to transfer
information from one point in the search tree (a dead end) to another.
It illustrates that Prolog is a multi-faceted language. On the one
hand it allows for pure logic programming, on the other hand it is a
very flexible tool for rapid prototyping.  Note that the savings due
to the reduction of the search space could be undone by the overhead
of computing and maintaining the extra information, especially, when
the amount of computation between two choice points is small.

The combination of look-back and look-ahead techniques can be useful,
and algorithms integrating both can be found, {\em e.g.},
\cite{Dechter02}. The question arises whether our solution can be
extended to incorporate look-ahead. This requires some work, however,
much of the design can be preserved. The initialization ({\tt
  init\_domains/3}) should not only associate variables with their initial
finite domain, but also with their eliminating explanations (initially
empty). Then the code for the main iteration could be as follows:
\begin{verbatim} 
extend_solution([],Solution,Solution).
extend_solution(Vars,PartialSolution,Solution) :-
        selectbestvar(Vars,var(Var,Values,Explanation),Rest),
        myassign(Values,Var,Explanation,Value),
        consistent(PartialSolution,assign(Var,Value)),
        propagate([assign(Var,Value)|PartialSolution],
                  NewPartialSolution)
        extend_solution(Vars,NewPartialSolution,Solution).
\end{verbatim} 

\noindent
The predicate {\tt selectbestvar/3} is used to dynamically select the
next variable to assign. It returns the identity of the variable
($\mathit{ Var}$), the available values ($\mathit{ Values}$) and the
explanation ($\mathit{ Explanation}$) for the eliminated values. When
a partial solution is successfully extended, the predicate {\tt
  propagate/2} has to take care of the constraint propagation:
eliminating values from domains and updating the corresponding
explanations after which the next iteration can start. Computing the
eliminating explanation for each eliminated value requires great care
and depends on the kind of look-ahead technique used. It is pretty
straightforward for forward checking but requires careful analysis in
case of {\em e.g.}, arc consistency as no pruning will occur on
backjumping when the elimination is attributed to {\em all} already
assigned variables.

\section*{Acknowledgments}

I am grateful to Bart Demoen, Gerda Janssens and Henk Vandecasteele
for useful comments on various drafts of this pearl. I am very
grateful to the reviewers. Indeed, as often is the case, their
persistence and good advise greatly contributed to the clarity of the
exposition.


\label{lastpage}

\end{document}